\documentclass[runningheads]{llncs}
\usepackage{graphicx}
\usepackage{amsmath,amssymb} 
\usepackage{color}
\usepackage{stmaryrd}
\usepackage{booktabs}

\begin{document}
\pagestyle{headings}
\mainmatter

\title{The Sound of Pixels} 

\titlerunning{The Sound of Pixels}

\authorrunning{Hang Zhao \textit{et al.}}

\author{Hang Zhao\inst{1} \and Chuang Gan\inst{1,2} \and Andrew Rouditchenko\inst{1} \and Carl Vondrick\inst{1,3}\and \\
Josh McDermott\inst{1} \and Antonio Torralba\inst{1}}

\institute{Massachusetts Institute of Technology\\
\and
MIT-IBM Watson AI Lab\\ 
\and
Columbia University \\
\email{\{hangzhao,roudi,jhm,torralba\}@mit.edu} \\
\email{\{ganchuang1990,cvondrick\}@gmail.com}
}

\maketitle

\begin{abstract}
We introduce PixelPlayer, a system that, by leveraging large amounts of unlabeled videos, learns to locate image regions which produce sounds and separate the input sounds into a set of components that represents the sound from each pixel.
Our approach capitalizes on the natural synchronization of the visual and audio modalities to learn models that jointly parse sounds and images, without requiring additional manual supervision. Experimental results  on a newly collected MUSIC dataset show that our proposed Mix-and-Separate framework outperforms several baselines on source separation. Qualitative results suggest our model learns to ground sounds in vision, enabling applications such as independently adjusting the volume of sound sources.


\end{abstract}

\begin{figure}[t]
\centering
\includegraphics[width=1\textwidth]{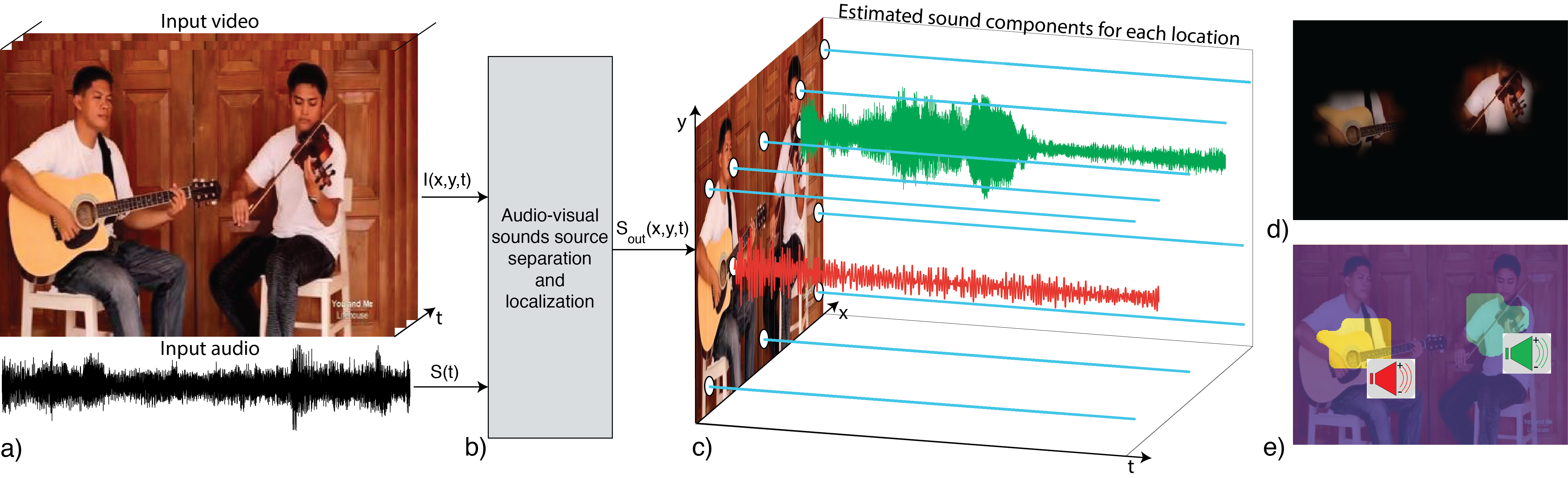}
\caption{PixelPlayer localizes sound sources in a video and separates the audio into its components without supervision. The figure shows: a) The input video frames $I(x,y,t)$, and the video mono sound signal $S(t)$. b) The system estimates the output sound signals $S_{out}(x,y,t)$ by separating the input sound. Each output component corresponds to the sound coming from a spatial location $(x,y)$ in the video. c) Component audio waveforms at 11 example locations; straight lines indicate silence. d) The system's estimation of the sound energy (or volume) of each pixel. e) Clustering of sound components in the pixel space. The same color is assigned to pixels with similar sounds. As an example application of clustering, PixelPlayer would enable the independent volume control of different sound sources in videos.}
\label{fig:teaser}
\end{figure}

\section{Introduction}

The world generates a rich source of visual and auditory signals. Our visual and auditory systems are able to recognize objects in the world, segment image regions covered by the objects, and isolate sounds produced by objects. While auditory scene analysis \cite{bregman1994auditory} is widely studied in the fields of environmental sound recognition \cite{mesaros2017dcase,hershey2017cnn} and source separation \cite{belouchrani1997blind,cardoso1997infomax,zibulevsky2001blind,vincent2006performance,virtanen2007monaural,comon2010handbook}, the natural synchronization between vision and sound can provide a rich supervisory signal for grounding sounds in vision \cite{Hershey1999,Kidron2005,Ngiam2011}. Training systems to recognize objects from vision or sound typically requires large amounts of supervision. In this paper, however, we leverage joint audio-visual learning to discover objects that produce sound in the world without manual supervision \cite{sa1993,owens2016visually,arandjelovic2017look}.

We show that by working with both auditory and visual information, we can learn in an unsupervised way to recognize objects from their visual appearance or the sound they make, to localize objects in images, and to separate the audio component coming from each object. We introduce a new system called PixelPlayer. Given an input video, PixelPlayer jointly separates the accompanying audio into components and spatially localizes them in the video. PixelPlayer enables us to listen to the sound originating from each pixel in the video. 

Fig.~\ref{fig:teaser} shows a working example of PixelPlayer (check the project website\footnote{http://sound-of-pixels.csail.mit.edu} for sample videos and interactive demos). In this example, the system has been trained with a large number of videos containing people playing instruments in different combinations, including solos and duets. No label is provided on what instruments are present in each video, where they are located, and how they sound. During test time, the input (Fig.~\ref{fig:teaser}.a) is a video of several instruments played together containing the visual frames $I(x,y,t)$, and the mono audio $S(t)$. PixelPlayer performs audio-visual source separation and localization, splitting the input sound signal to estimate output sound components $S_{out}(x,y,t)$, each one corresponding to the sound coming from a spatial location $(x,y)$ in the video frame. As an illustration, Fig.~\ref{fig:teaser}.c shows the recovered audio signals for 11 example pixels. The flat blue lines correspond to pixels that are considered as silent by the system. The non-silent signals correspond to the sounds coming from each individual instrument. Fig.~\ref{fig:teaser}.d shows the estimated sound energy, or volume of the audio signal from each pixel. Note that the system correctly detects that the sounds are coming from the two instruments and not from the background. Fig.~\ref{fig:teaser}.e shows how pixels are clustered according to their component sound signals. The same color is assigned to pixels that generate very similar sounds. 

The capability to incorporate sound into vision will have a large impact on a range of applications involving the recognition and manipulation of video. PixelPlayer's ability to separate and locate sounds sources will allow more isolated processing of the sound coming from each object and will aid auditory recognition. Our system could also facilitate sound editing in videos, enabling, for instance, volume adjustments for specific objects or removal of the audio from particular sources.

Concurrent to this work, there are papers \cite{ephrat2018looking,owens2018audio} at the same conference that also show the power of combining vision and audio to decompose sounds into components. \cite{ephrat2018looking} shows how person appearance could help solving the cocktail party problem in speech domain.
\cite{owens2018audio} demonstrates an audio-visual system that separates on-screen sound \textit{vs.} background sounds not visible in the video.

This paper is presented as follows. In Section \ref{sec:related}, we first review related work in both the vision and sound communities. In Section \ref{sec:system}, we present our system that leverages cross-modal context as a supervisory signal. In Section \ref{sec:dataset}, we describe a new dataset for visual-audio grounding. In Section \ref{sec:experiment}, we present several experiments to analyze our model. Subjective evaluations are presented in Section \ref{sec:sub_eval}.

\section{Related Work}
\label{sec:related}
Our work relates mainly to the fields of sound source separation, visual-audio cross-modal learning, and self-supervised learning, which will be briefly discussed in this section. 

\textbf{Sound source separation.}
Sound source separation, also known as the ``cocktail party problem"~\cite{mcdermott2009cocktail,haykin2005cocktail}, is a classic problem in engineering and perception. Classical approaches include signal processing methods such as Non-negative Matrix Factorization (NMF) \cite{virtanen2007monaural,cichocki2009nonnegative,smaragdis2003non}. More recently, deep learning methods have gained popularity \cite{wang2017supervised,chandna2017monoaural}. Sound source separation methods enable applications ranging from music/vocal separation \cite{simpson2015deep}, to speech separation and enhancement \cite{hershey2016deep,gabbay2017seeing,nagrani2018seeing}. Our problem differs from classic sound source separation problems because we want to separate sounds into visually and spatially grounded components. 

\textbf{Learning visual-audio correspondence.} Recent work in computer vision has explored the relationship between vision and sound. One line of work has developed models for generating sound from silent videos \cite{owens2016visually,zhou2017visual}. The correspondence between vision and sound has also been leveraged for learning representations. For example, \cite{owens2016ambient} used audio to supervise visual representations, \cite{aytar2016soundnet,hershey2017cnn} used vision to supervise audio representations, and \cite{arandjelovic2017look} used sound and vision to jointly supervise each other. In work related to our paper, people studied how to localize sounds in vision according to motion~\cite{izadinia2013multimodal} or semantic cues~\cite{arandjelovic2017objects,senocak2018learning}, however they do not separate multiple sounds from a mixed signal. 

\textbf{Self-supervised learning.} Our work builds off efforts to learn perceptual models that are ``self-supervised'' by leveraging natural contextual signals in images \cite{doersch2015unsupervised,larsson2017colorization,pathak2016context,shu2017neural,ma2018single}, videos \cite{wang2015unsupervised,pathak2017learning,vondrick2016generating,vondrick2018tracking,Gan2018geometry,jayaraman2015learning}, and even radio signals~\cite{zhao2018through}. These approaches utilize the power of supervised learning while not requiring manual annotations, instead deriving supervisory signals from the structure in natural data. Our model is similarly self-supervised, but uses self-supervision to learn to separate and ground sound in vision.

\begin{figure}[t]
\centering
\includegraphics[width=1\textwidth]{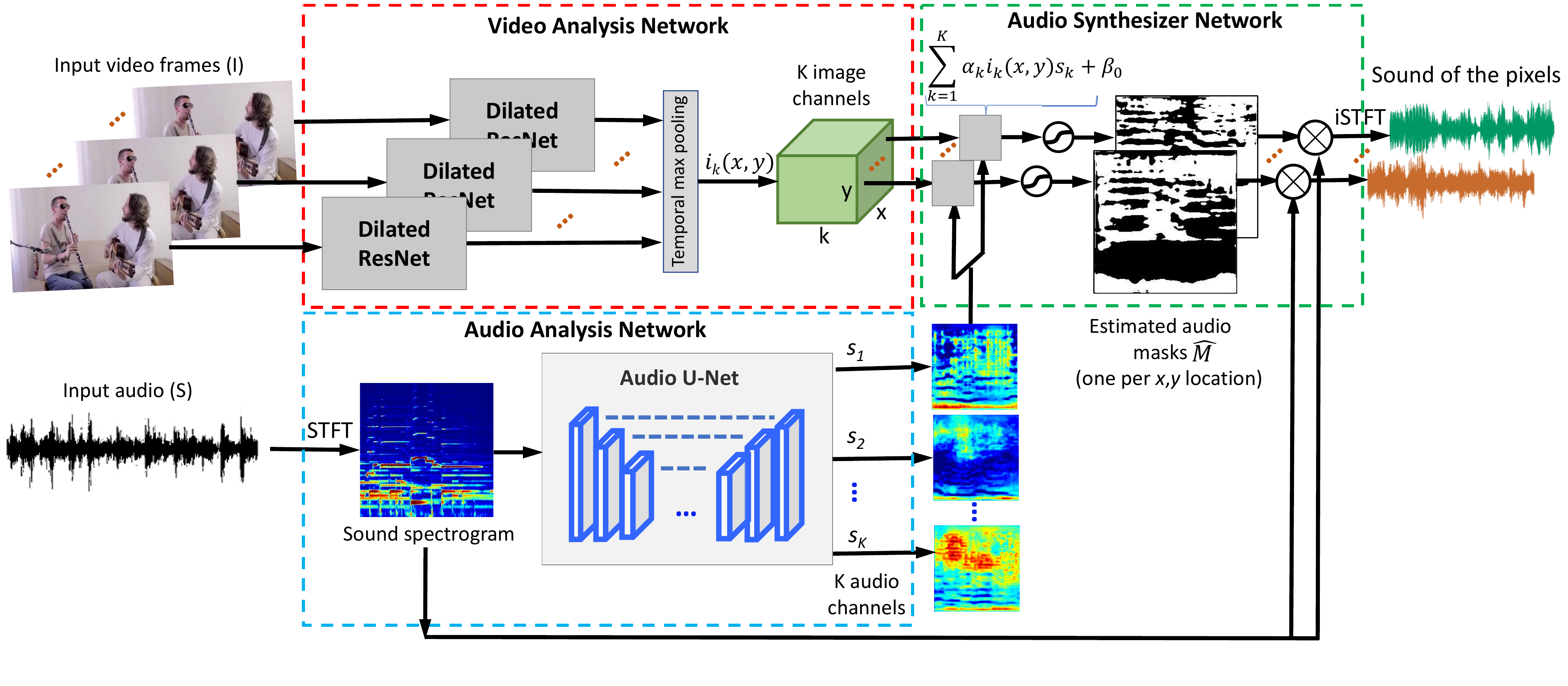}
\vspace{-0.3in}
\caption{Procedure to generate the sound of a pixel: pixel-level visual features are extracted by temporal max-pooling over the output of a dilated ResNet applied to T frames. The input audio spectrogram is passed through a U-Net whose output is K audio channels. The sound of each pixel is computed by an audio synthesizer network. The audio synthesizer network outputs a mask to be applied to the input spectrogram that will select the spectral components associated with the pixel. Finally, inverse STFT is applied to the spectrogram computed for each pixel to produce the final sound.}
\label{fig:model_test2}
\end{figure}

\section{Audio-Visual Source Separation and Localization}
\label{sec:system}
\vspace{-0.1in}
In this section, we introduce the model architectures of PixelPlayer, and the proposed Mix-and-Separate training framework that learns to separate sound according to vision. 
\vspace{-0.1in}
\subsection{Model architectures}
Our model is composed of a video analysis network, an audio analysis network, and an audio synthesizer network, as shown in Fig.~\ref{fig:model_test2}. 

\begin{figure}[t]
\centering
\includegraphics[width=1.0\textwidth]{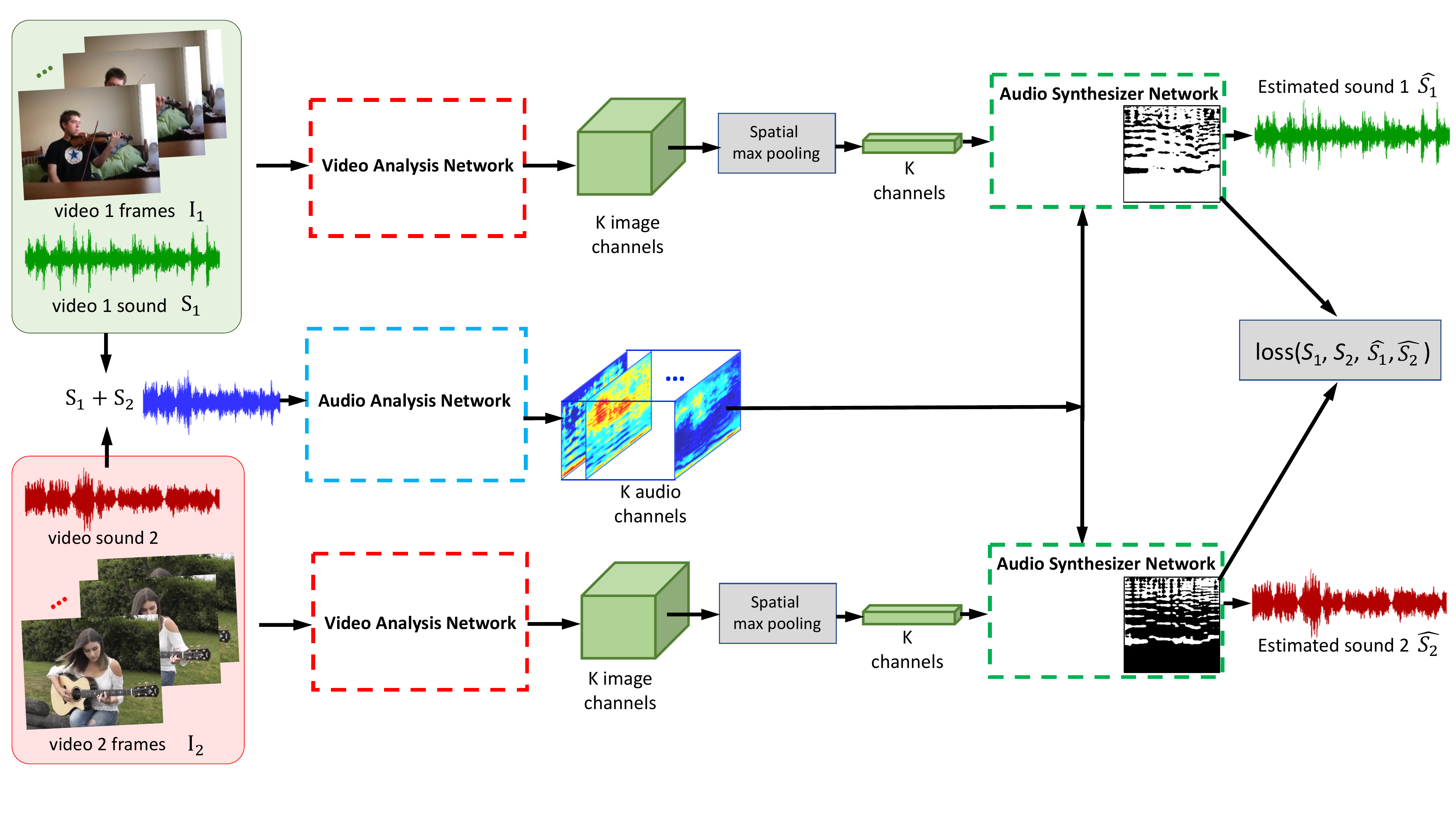}
\vspace{-0.3in}
\caption{Training pipeline of our proposed Mix-and-Separate framework in the case of mixing two videos ($N=2$). The dashed boxes represent the modules detailed in Fig.~\ref{fig:model_test2}. The audio signals from the two videos are added together to generate an input mixture with known constituent source signals. The network is trained to separate the audio source signals conditioned on corresponding video frames; its output is an estimate of both sound signals. Note that we do not assume that each video contains a single source of sound. Moreover, no annotations are provided. The system thus learns to separate individual sources without traditional supervision.}
\label{fig:model_train}
\vspace{-0.1in}
\end{figure}

\vspace{-0.1in}
\subsubsection{Video analysis network.} The video analysis network extracts visual features from video frames. Its choice can be an arbitrary architecture used for visual classification tasks. Here we use a dilated variation of the ResNet-18 model \cite{he2016deep} which will be described in detail in the experiment section. For an input video of size T$\times$H$\times$W$\times$3, the ResNet model extracts per-frame features with size \newline T$\times$(H/16)$\times$(W/16)$\times$K. 
After \texttt{temporal pooling} and \texttt{sigmoid} activation, we obtain a visual feature $i_k(x,y)$ for each pixel with size K. 

\vspace{-0.1in}
\subsubsection{Audio analysis network.} The audio analysis network takes the form of a U-Net \cite{ronneberger2015u} architecture, which splits the input sound into K components $s_k$, $k=(1,...,K)$. We empirically found that working with audio spectrograms gives better performance than using raw waveforms, so the network described in this paper uses the Time-Frequency (T-F) representation of sound. First, a Short-Time Fourier Transform (STFT) is applied on the input mixture sound to obtain its spectrogram. Then the magnitude of spectrogram is transformed into log-frequency scale (analyzed in Sec.~\ref{sec:experiment}), and fed into the U-Net which yields K feature maps containing features of different components of the input sound.

\subsubsection{Audio synthesizer network.} The synthesizer network finally predicts the predicted sound by taking pixel-level visual feature $i_k(x,y)$ and audio feature $s_k$. The output sound spectrogram is generated by vision-based spectrogram masking technique. Specifically, a mask $M(x,y)$ that could separate the sound of the pixel from the input is estimated, and multiplied with the input spectrogram. Finally, to get the waveform of the prediction, we combine the predicted magnitude of spectrogram with the phase of input spectrogram, and use inverse STFT for recovery.


\subsection{Mix-and-Separate framework for Self-supervised Training}
The idea of the Mix-and-Separate training procedure is to artificially create a complex auditory scene and then solve the auditory scene analysis problem of separating and grounding sounds.
Leveraging the fact that audio signals are approximately additive, we mix sounds from different videos to generate a complex audio input signal.
The learning objective of the model is to separate a sound source of interest conditioned on the visual input associated with it. 

Concretely, to generate a complex audio input, we randomly sample $N$ videos $\{I_n, S_n\}$ from the training dataset, where $n=(1,...,N)$. $I_n$ and $S_n$ represent the visual frames and audio of the $n$-th video, respectively. The input sound mixture is created through linear combinations of the audio inputs as $S_{mix} = \sum_{n=1}^N S_n$. The model $f$ learns to estimate the sounds in each video $\hat{S_n}$ given the audio mixture and the visual of the corresponding video $\hat{S_n} = f(S_{mix}, I_n)$.

Fig.~\ref{fig:model_train} shows the training framework in the case of $N=2$. The training phase differs from the testing phase in that 1) we sample multiple videos randomly from the training set, mix the sample audios and target to recover each of them given their corresponding visual input; 2) video-level visual features are obtained by \texttt{spatial-temporal max pooling} instead of pixel-level features. Note that although we have clear targets to learn in the training process, it is still unsupervised as we do not use the data labels and do not make assumptions about the sampled data.

The learning target in our system are the spectrogram masks, they can be binary or ratios. In the case of binary masks, the value of the ground truth mask of the $n$-th video is calculated by observing whether the target sound is the dominant component in the mixed sound in each T-F unit, 
\begin{equation}
	M_n(u, v) = \llbracket S_n(u, v) \ge S_m(u, v)\rrbracket, \quad \forall m=(1,...,N),
\end{equation}
where $(u, v)$ represents the coordinates in the T-F representation and $S$ represents the spectrogram. Per-pixel sigmoid cross entropy loss is used for learning. For ratio masks, the ground truth mask of a video is calculated as the ratio of the magnitudes of the target sound and the mixed sound,
\begin{equation}
	M_n(u, v) = \frac{S_n(u, v)}  {S_{mix}(u, v)}.
\end{equation}
In this case, per-pixel $L1$ loss~\cite{zhao2017loss} is used for training.
Note that the values of the ground truth mask do not necessarily stay within $[0, 1]$ because of interference.

\section{MUSIC Dataset}
\label{sec:dataset}
The most commonly used videos with audio-visual correspondence are musical recordings, so we introduce a musical instrument video dataset for the proposed task, called MUSIC (Multimodal Sources of Instrument Combinations) dataset.

\begin{figure}[t]
\centering
\includegraphics[width=1\textwidth]{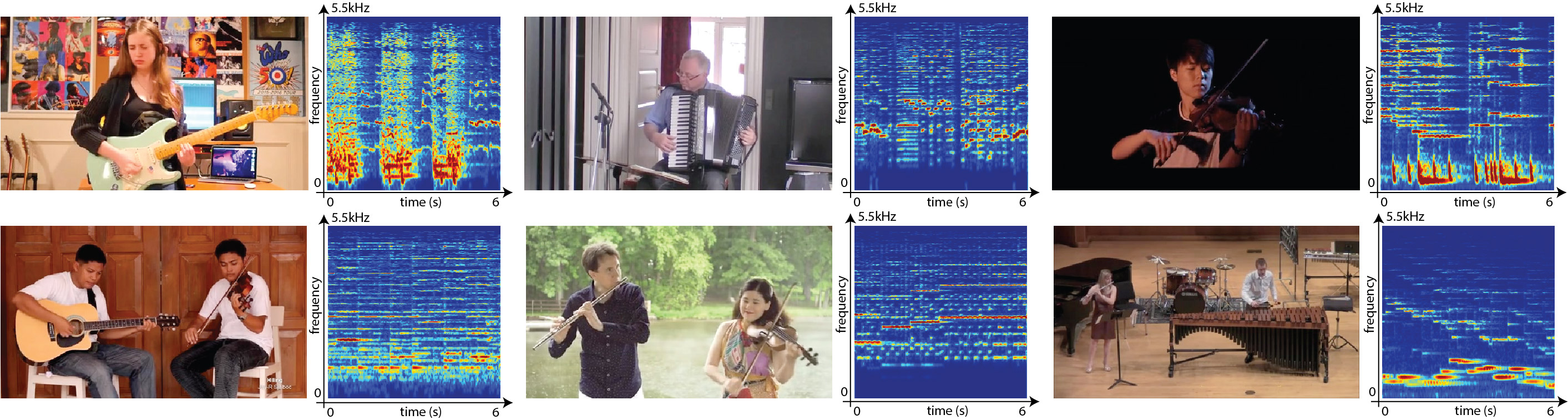}
\caption{Example frames and associated sounds from our video dataset. The top row shows videos of solos and the bottom row shows videos of duets. The sounds are displayed in the time-frequency domain as spectrograms, with frequency on a log scale. }
\label{fig:dataset}
\end{figure}

\begin{figure}[t]
\centering
\includegraphics[width=1.0\textwidth]{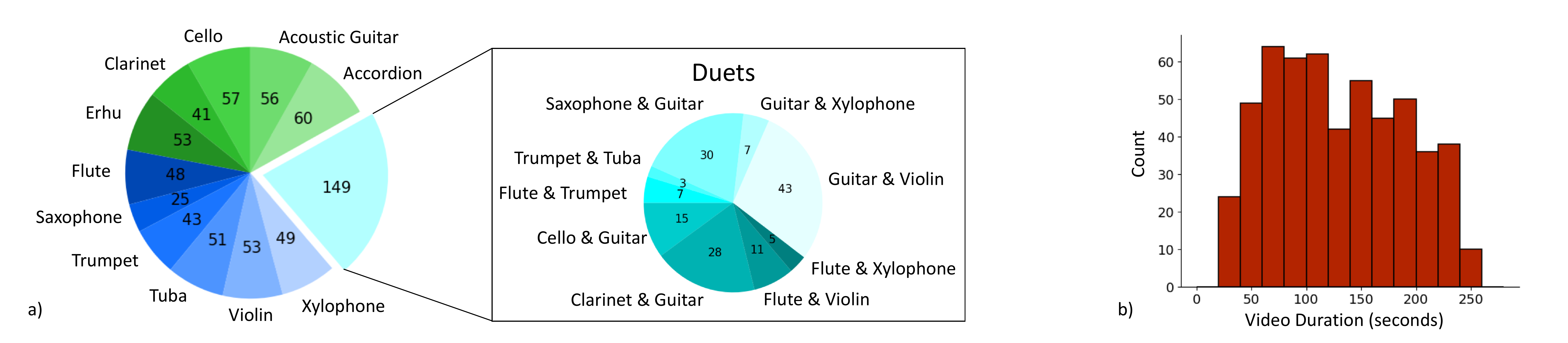}
\caption{Dataset Statistics: a) Shows the distribution of video categories. There are 536 videos of solos and 149 videos of duets. b) Shows the distribution of the solo video durations. The average duration is about 2 minutes.  }
\label{fig:dataset_stats}
\vspace{-0.1in}
\end{figure}

We retrieved the MUSIC videos from YouTube by keyword query. During the search, we added keywords such as ``cover'' to find more videos that were not post-processed or edited.

MUSIC dataset has 685 untrimmed videos of musical solos and duets, some sample videos are shown in Fig.~\ref{fig:dataset}. The dataset spans 11 instrument categories: accordion, acoustic guitar, cello, clarinet, erhu, flute, saxophone, trumpet, tuba, violin and xylophone. Fig.~\ref{fig:dataset_stats} shows the dataset statistics. 

Statistics reveal that due to the natural distribution of videos, duet performances are less balanced than the solo performances. For example, there are almost no videos of tuba and violin duets, while there are many videos of guitar and violin duets.

\section{Experiments}
\label{sec:experiment}
\subsection{Audio data processing}
There are several steps we take before feeding the audio data into our model. To speed up computation, we sub-sampled the audio signals to 11kHz, such that the highest signal frequency preserved is 5.5kHz. This preserves the most perceptually important frequencies of instruments and only slightly degrades the overall audio quality.
Each audio sample is approximately 6 seconds, randomly cropped from the untrimmed videos during training. An STFT with a window size of 1022 and a hop length of 256 is computed on the audio samples, resulting in a $512\times256$ Time-Frequency (T-F) representation of the sound. We further re-sample this signal on a log-frequency scale to obtain a $256\times256$ T-F representation. This step is similar to the common practice of using a Mel-Frequency scale, \textit{e.g.} in speech recognition \cite{logan2000mel}. The log-frequency scale has the dual advantages of (1) similarity to the frequency decomposition of the human auditory system (frequency discrimination is better in absolute terms at low frequencies) and (2) translation invariance for harmonic sounds such as musical instruments (whose fundamental frequency and higher order harmonics translate on the log-frequency scale as the pitch changes), fitting well to a ConvNet framework. 
The log magnitude values of T-F units are used as the input to the audio analysis network. After obtaining the output mask from our model, we use an inverse sampling step to convert our mask back to linear frequency scale with size $512\times256$, which can be applied on the input spectrogram. We finally perform an inverse STFT to obtain the recovered signal.

\subsection{Model configurations}
In all the experiments, we use a variant of the ResNet-18 model for the video analysis network, with the following modifications made: (1) removing the last \texttt{average pooling} layer and \texttt{fc} layer; (2) removing the stride of the last residual block, and making the convolution layers in this block to have a dilation of $2$; (3) adding a last $3\times3$ convolution layer with K output channels. For each video sample, it takes T frames with size $224\times224\times3$ as input, and outputs a feature of size K after \texttt{spatiotemporal max pooling}.

The audio analysis network is modified from U-Net. It has 7 convolutions (or down-convolutions) and 7 de-convolutions (or up-convolution) with skip connections in between.
It takes an audio spectrogram with size $256\times256\times1$, and outputs K feature maps of size $256\times256\times K$.

The audio synthesizer takes the outputs from video and audio analysis networks, fuses them with a weighted summation, and outputs a mask that will be applied on the spectrogram. The audio synthesizer is a linear layer which has very few trainable parameters (K weights + 1 bias). It could be designed to have more complex computations, but we choose the simple operation in this work to show interpretable intermediate representations, which will be shown in Sec \ref{sec:activation}.

Our best model takes 3 frames as visual input, and uses the number of feature channels $K=16$.

\subsection{Implementation details}
Our goal in the model training is to learn on natural videos (with both solos and duets), evaluate quantitatively on the validation set, and finally solve the source separation and localization problem on the natural videos with mixtures. Therefore, we split our MUSIC dataset into 500 videos for training, 130 videos for validation, and 84 videos for testing. Among them, 500 training videos contain both solos and duets, the validation set only contains solos, and the test set only contains duets.

During training, we randomly sample $N=2$ videos from our MUSIC dataset, which can be solos, duets, or silent background. Silent videos are made by pairing silent audio waveforms randomly with images from the ADE dataset \cite{zhou2017scene} which contains images of natural environments. This technique regularizes the model better in localizing objects that sound by introducing more silent videos. To recap, the input audio mixture could contain 0 to 4 instruments. We also experimented with combining more sounds, but that made the task more challenging and the model did not learn better. 

In the optimization process, we use a SGD optimizer with momentum $0.9$. We set the learning rate of the audio analysis network and the audio synthesizer both as $0.001$, and the learning rate of the video analysis network as $0.0001$ since we adopt a pre-trained CNN model on ImageNet.

\subsection{Sound Separation Performance}
\label{sec:eval_separation}
To evaluate the performance of our model, we also use the Mix-and-Separate process to make a validation set of synthetic mixture audios and the separation is evaluated.

\begin{figure}[t]
\centering
\includegraphics[width=0.95\textwidth]{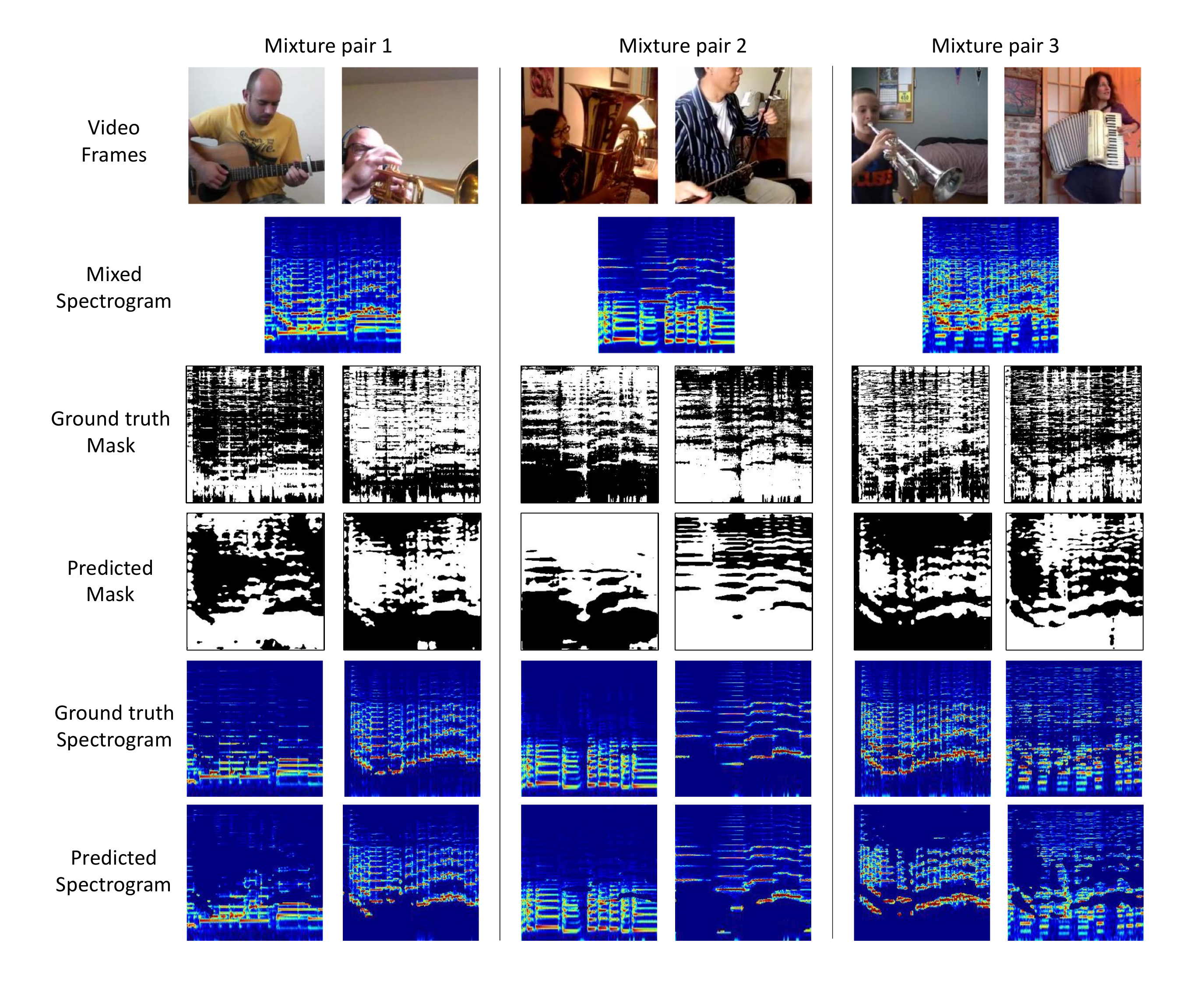}
\vspace{-0.3in}
\caption{Qualitative results on vision-guided source separation on synthetic audio mixtures. This experiment is performed only for quantitative model evaluation.}
\label{fig:result_spectrogram}
\vspace{-10px}
\end{figure}

Fig. \ref{fig:result_spectrogram} shows qualitative results of our best model, which predicts binary masks that apply on the mixture spectrogram. The first row shows one frame per sampled videos that we mix together, the second row shows the spectrogram (in log frequency scale) of the audio mixture, which is the actual input to the audio analysis network. The third and fourth rows show ground truth masks and the predicted masks, which are the targets and output of our model. The fifth and sixth rows show the ground truth spectrogram and predicted spectrogram after applying masks on the input spectrogram. We could observe that even with the complex patterns in the mixed spectrogram, our model can ``segment'' the target instrument components out successfully.


\begin{table}[t]
	\begin{center}
	\begin{tabular}{l|| c | c | c | c|c | c|c}
    & NMF & DeepConvSep & Spectral &  \multicolumn{2}{c|}{Ratio Mask}  & \multicolumn{2}{c}{Binary Mask}  \\
    & \cite{virtanen2007monaural} & \cite{chandna2017monoaural} & Regression  & Linear scale  & Log scale & Linear scale & Log scale \\ \hline \hline
   NSDR & 3.14   & 6.12 &  5.12     &  6.67        & 8.56     &  6.94     & \textbf{8.87} \\ \hline
   SIR  & 6.70   & 8.38 &  7.72     &  12.85        & 13.75      &  12.87        & \textbf{15.02}  \\ \hline
   SAR  & 10.10  & 11.02 &  10.43    &  13.87       &  \textbf{14.19}&   11.12     & 12.28  \\ \hline
	\end{tabular}
	\end{center}
	\caption{Model performances of baselines and different variations of our proposed model, evaluated in NSDR/SIR/SAR. Binary masking in log frequency scale performs best in most metrics.}

\label{tab:obj_eval}
\end{table}

To quantify the performance of the proposed model, we use the following metrics: the Normalized Signal-to-Distortion Ratio (NSDR), Signal-to-Interference Ratio (SIR), and Signal-to-Artifact Ratio (SAR) on the validation set of our synthetic videos. The NSDR is defined as the difference in SDR of the separated signals compared with the ground truth signals and the SDR of the mixture signals compared with the ground truth signals. This represents the improvement of using the separated signal compared with using the mixture as each separated source. The results reported in this paper were obtained by using the open-source \texttt{mir\_eval}  \cite{raffel2014mir_eval} library. 

Results are shown in Table~\ref{tab:obj_eval}. Among all the models, baseline approaches NMF~\cite{virtanen2007monaural} and DeepConvSep~\cite{chandna2017monoaural} use audio and ground-truth labels to do source separation. All variants of our model use the same architecture we described, and take both visual and sound input for learning. Spectral Regression refers to the model that directly regresses output spectrogram values given an input mixture spectrogram, instead of outputting spectrogram mask values. From the numbers in the table, we can conclude that (1) masking based approaches are generally better than direct regression; (2) working in the log frequency scale performs better than in the linear frequency scale; (3) binary masking based method achieves similar performance as ratio masking.

Meanwhile, we found that the NSDR/SIR/SAR metrics are not the best metrics for evaluating perceptual separation quality, so in Sec~\ref{sec:sub_eval} we further conduct user studies on the audio separation quality.

\vspace{-5px}
\subsection{Visual Grounding of Sounds}
As the title of paper indicates, we are fundamentally solving two problems: localization and separation of sounds. 

\vspace{-5px}
\subsubsection{Sound localization.}
\begin{figure}[t]
\centering
\includegraphics[width=0.95\textwidth]{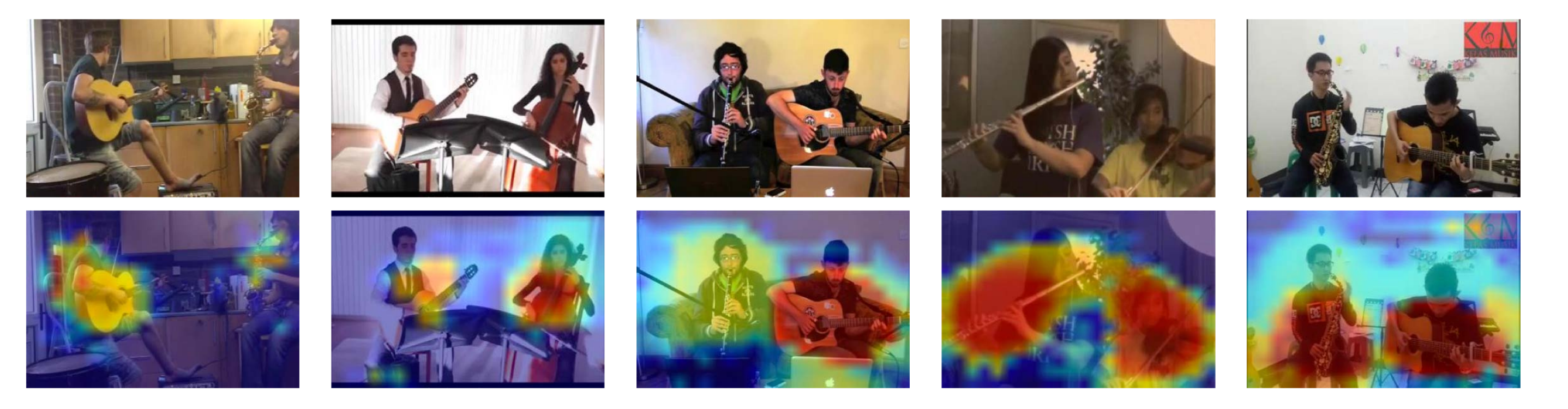}
\vspace{-15px}
\caption{``Which pixels are making sounds?" Energy distribution of sound in pixel space. Overlaid heatmaps show the volumes from each pixel.}
\label{fig:result_energy}
\end{figure}

The first problem is related to the spatial grounding question, ``which pixels are making sounds?'' This is answered in Fig.~\ref{fig:result_energy}: for natural videos in the dataset, we calculate the sound energy (or volume) of each pixel in the image, and plot their distributions in heatmaps. As can be seen, the model accurately localizes the sounding instruments.

\vspace{-5px}
\subsubsection{Clustering of sounds.}
\begin{figure}[t]
\centering
\includegraphics[width=0.95\textwidth]{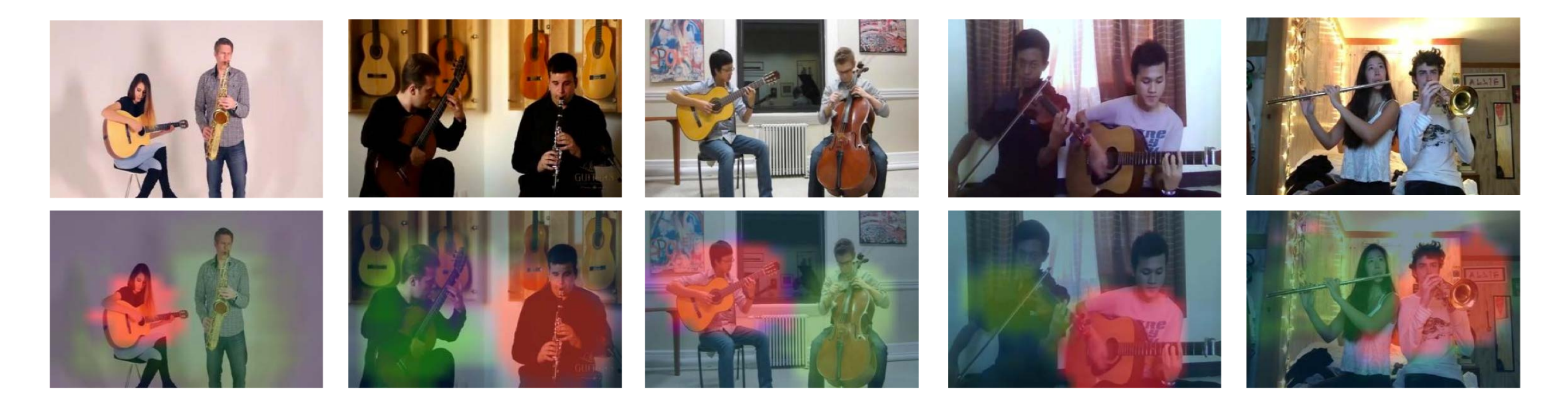}
\vspace{-15px}
\caption{``What sounds do these pixels make?" Clustering of sound in space. Overlaid colormap shows different audio features with different colors.}
\label{fig:result_clustering}
\vspace{-10px}
\end{figure}

The second problem is related to a further question: ``what sounds do these pixels make?'' In order to answer this, we visualize the sound each pixel makes in images in the following way: for each pixel in a video frame, we take the feature of its sound, namely the vectorized log spectrogram magnitudes, and project them onto 3D RGB space using PCA for visualization purposes. Results are shown in Fig.~\ref{fig:result_clustering}, different instruments and the background in the same video frame have different color embeddings, indicating different sounds that they make.

\subsubsection{Discriminative channel activations.}
\label{sec:categorization}
\begin{figure}[t]
\centering
\includegraphics[width=0.8\textwidth]{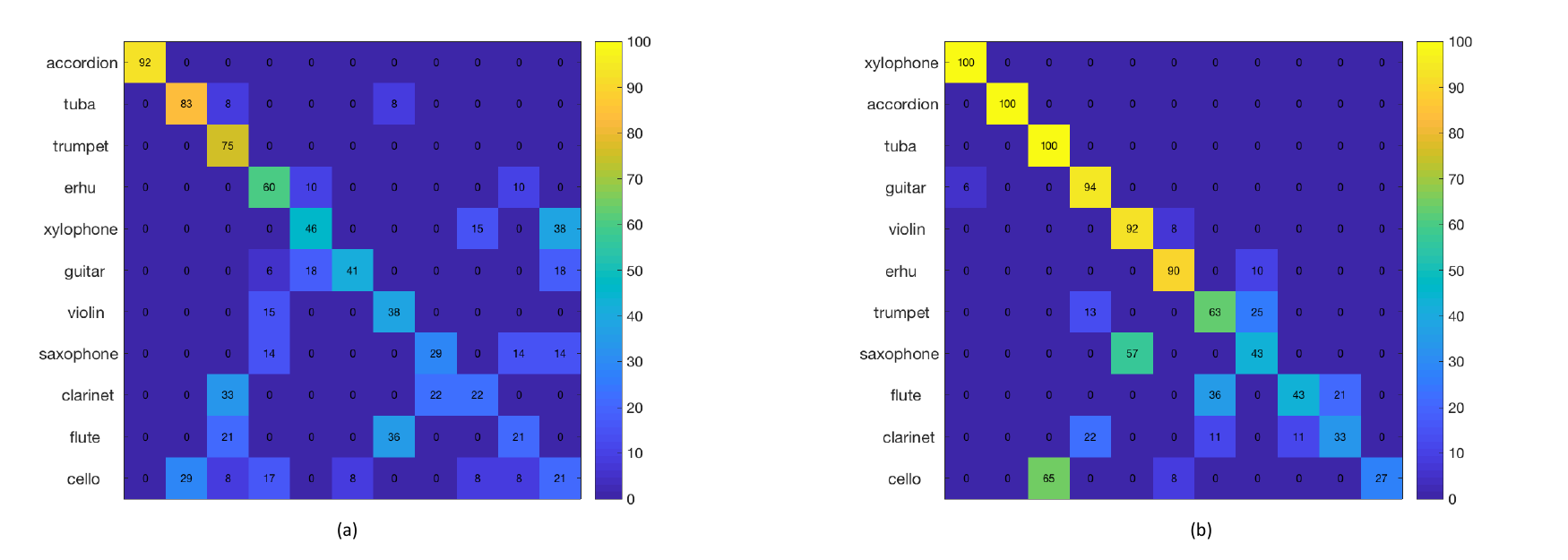}
\vspace{-0.1in}
\caption{(a) Visual and (b) audio confusion matrices by sorting channel activations with respect to ground truth category labels. }
\label{fig:confusion}
\end{figure}

Given our model could separate sounds of different instruments, we explore its channel activations for different categories. For validation samples of each category, we find the strongest activated channel, and then sort them to generate a confusion matrix. Fig.~\ref{fig:confusion} shows the (a) visual and (b) audio confusion matrices from our best model. If we simply evaluate classification by assigning one category to one channel, the accuracy is $46.2\%$ for vision and $68.9\%$ for audio. Note that no learning is involved here, we expect much higher performance by using a linear classifier. This experiment demonstrates that the model has implicitly learned to discriminate instruments visually and auditorily. 

In a similar fashion, we evaluate object localization performance of the video analysis network based on the channel activations. To generate a bounding box from the channel activation map, we follow \cite{zhou2016learning} to threshold the map. We first segment the regions of which the value is above 20\% of the max value of the activation map, and then take the bounding box that covers the largest connected component in the segmentation map. Localization accuracy under different intersection over union (IoU) criterion are shown in Table \ref{tab:localization}.

\begin{table}[t]
	\begin{center}
	\begin{tabular}{l||c|c|c}
     IoU Threshold~~&~~~~0.3~~~~&~~~~0.4~~~~&~~~~0.5~~~~\\ \hline
   	 Accuracy(\%) & 66.10   & 47.92  & 32.43   \\ \hline
	\end{tabular}
	\end{center}
	\caption{Object localization performance of the learned video analysis network.}

\label{tab:localization}
\vspace{-15px}
\end{table}

\subsection{Visual-audio corresponding activations}
\label{sec:activation}
\begin{figure}[t]
\centering
\includegraphics[width=1\textwidth]{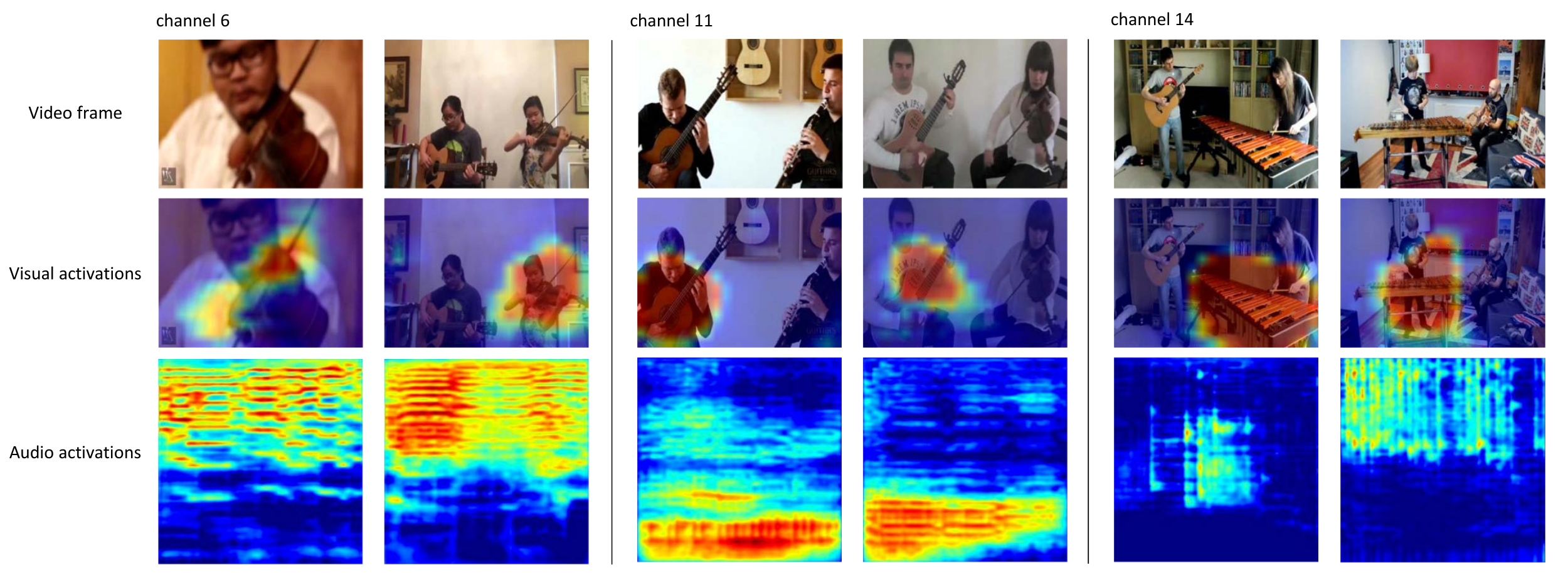}
\caption{Visualizations of corresponding channel activations. Channel 6 has emerged as a violin detector, responding strongly to the presence of violins in the video frames and to the high order harmonics in the spectrogram, which are colored brighter in the spectrogram of the figure. Likewise, channel 11 and 14 seems to detect the visual and auditory characteristics of guitars and xylophones. }
\label{fig:result_channels}
\end{figure}

As our proposed model is a form of self-supervised learning and is designed such that both visual and audio networks learn to activate simultaneously on the same channel, we further explore the representations learned by the model. Specifically, we look at the K channel activations of the video analysis network before \texttt{max pooling}, and their corresponding channel activations of the audio analysis network. The model has learned to detect important features of specific objects across the individual channels. In Fig.~\ref{fig:result_channels} we show the top activated videos of channel 6, 11 and 14. These channels have emerged as violin, guitar and xylophone detectors respectively, in both visual and audio domains. Channel 6 responds strongly to the visual appearance of violin and to the higher order harmonics in violin sounds. Channel 11 responds to guitars and the low frequency region in sounds. And channel 14 responds to the visual appearance of xylophone and to the brief, pulse-like patterns in the spectrogram domain. For other channels, some of them also detect specific instruments while others just detect specific features of instruments.

\section{Subjective Evaluations}
\label{sec:sub_eval}

The objective and quantitative evaluations in Sec.~\ref{sec:eval_separation} are mainly performed on the synthetic mixture videos, the performance on the natural videos needs to be further investigated. On the other hand, the popular NSDR/SIR/SAR metrics used are not closely related to perceptual quality. Therefore we conducted crowd-sourced subjective evaluations as a complementary evaluation. Two studies are conducted on Amazon Mechanical Turk (AMT) by human raters, a sound separation quality evaluation and a visual-audio correspondence evaluation. 

\subsection{Sound separation quality}
For the sound separation evaluation, we used a subset of the solos from the dataset as ground truth. We prepared the outputs of the baseline NMF model and the outputs of our models, including spectral regression, ratio masking and binary masking, all in log frequency scale. For each model, we take 256 audio outputs from the same set for evaluation and each audio is evaluated by 3 independent AMT workers. Audio samples are randomly presented to the workers, and the following question is asked: ``\texttt{Which sound do you hear? 1. A, 2. B, 3. Both, or 4. None of them}". Here \texttt{A} and \texttt{B} are replaced by their mixture sources, \textit{e.g.} \texttt{A}=clarinet, \texttt{B}=flute. 

Subjective evaluation results are shown in Table~\ref{tab:sub_eval_separation}. We show the percentages of workers who heard only the correct solo instrument (\texttt{Correct}), who heard only the incorrect solo instrument (\texttt{Wrong}), who heard both of the instruments (\texttt{Both}), and who heard neither of the instruments (\texttt{None}). First, we observe that although the NMF baseline did not have good NSDR numbers in the quantitative evaluation, it has competitive results in our human study. Second, among our models, the binary masking model outperforms all other models by a margin, showing its advantage in separation as a classification model. The binary masking model gives the highest correct rate, lowest error rate, and lowest confusion (percentage of \texttt{Both}), indicating that the binary model performs source separation perceptively better than the other models. It is worth noticing that even the ground truth solos do not give 100\% correct rate, which represents the upper bound of performance.

\begin{table}[t]
	\begin{center}	
    \begin{tabular}{c|c|c|c|c}
    	\textbf{Model} & \textbf{Correct}(\%) & \textbf{Wrong}(\%)& \textbf{Both}(\%) & \textbf{None}(\%) \\ 
        \hline \hline
        NMF & 45.70 &15.23 &21.35& 17.71\\ 
        Spectral Regression & 18.23 &15.36 & 64.45 & 1.95\\
        Ratio Mask & 39.19 & 19.53& 27.73 & 13.54\\
        Binary Mask & \textbf{59.11} & 11.59& 18.10 & 11.20 \\ 
        \hline
        Ground Truth Solo  & 70.31 & 16.02 & 7.68 & 5.99\\ \hline

	\end{tabular}
	\end{center}
	\caption{Subjective evaluation of sound separation performance. Binary masking-based model outperforms other models in sound separation.}

\label{tab:sub_eval_separation}
\end{table}

\subsection{Visual-sound correspondence evaluations}
The second study focuses on the evaluation of the visual-sound correspondence problem. For a pixel-sound pair, we ask the binary question: ``\texttt{Is the sound coming from this pixel?}''
For this task, we only evaluate our models for comparison as the task requires visual input, so audio-only baselines are not applicable.
We select 256 pixel positions (50\% on instruments and 50\% on background objects) to generate corresponding sounds with different models, and get the percentage of \texttt{Yes} responses from the workers, which tells the percentage of pixels with good source separation and localization, results are shown in Table~\ref{tab:sub_eval_correspondece}. This evaluation also demonstrates that the binary masking-based model gives the best performance in the vision-related source separation problem. 

\begin{table}[t]
	\begin{center}	
	\begin{tabular}{c|c}
    	\textbf{Model} & \textbf{Yes}(\%) \\ \hline \hline
        Spectral Regression & 39.06 \\ \hline
        Ratio Mask & 54.68 \\ \hline
        Binary Mask & 67.58 \\ \hline
	\end{tabular}
	\end{center}
	\caption{Subjective evaluation of visual-sound correspondence. Binary masking-based model best relates vision and sound.}
\label{tab:sub_eval_correspondece}
\end{table}

\section{Conclusions}

In  this paper, we introduced PixelPlayer, a system that learns from unlabeled videos to separate input sounds and also locate them in the visual input.
Quantitative results, qualitative results, and subjective user studies demonstrate the effectiveness of our cross-modal learning system. We expect our work can open up new research avenues for understanding the problem of sound source separation using both visual and auditory signals. 

\vspace{15px}
\noindent\textbf{Acknowledgement:} This work was supported by NSF grant IIS-1524817. We thank Adria Recasens, Yu Zhang and Xue Feng for insightful discussions.

\bibliographystyle{splncs04}
\bibliography{egbib}

\end{document}